%% file: lawsumm.tex
\documentclass[review]{elsarticle}

\usepackage{lineno,hyperref}
\usepackage{array,multirow,graphicx}
\usepackage{float}
\usepackage{arydshln}
\usepackage{geometry}

\modulolinenumbers[5]

\journal{Journal of Information Processing and Management}

\makeatletter
\def\ps@pprintTitle{%
  \let\@oddhead\@empty
  \let\@evenhead\@empty
  \let\@oddfoot\@empty
  \let\@evenfoot\@oddfoot
}
\makeatother

\begin{document}

\begin{frontmatter}

\title{\textbf{LawSum: A weakly supervised approach for Indian Legal
Document Summarization}}

\author[1]{\textbf{Vedant Parikh}}
\author[3]{\textbf{Vidit Mathur}}
\author[2]{\textbf{Parth Mehta}}
\author[3]{\textbf{Namita Mittal}}
\author[1]{\textbf{Prasenjit Majumder}}

\address[1]{Dhirubhai Ambani Institute of Information and Communication Technology}
\address[2]{Paramonic AI}
\address[3]{Malaviya National Institute of Technology}







\input{sections/abstract}


\input{sections/keywords}

\end{frontmatter}

\input{sections/introduction}
\input{sections/Related_Works}
\input{sections/dataset}
\input{sections/experiments}
\input{sections/Different_Approaches}
\input{sections/conclusion}



\section*{References}

\bibliographystyle{model1-num-names}
\bibliography{lawsumm}
\end{document}

%% file: sections/abstract.tex
\begin{abstract}
Unlike the courts in western countries, public records of Indian judiciary are completely unstructured and noisy. No large scale publicly available annotated datasets of Indian legal documents exist till date. This limits the scope for legal analytics research. In this work, we propose a new dataset consisting of over 10,000 judgements delivered by the supreme court of India and their corresponding hand written summaries. The proposed dataset is pre-processed by normalizing common legal abbreviations, handling spelling variations in named entities, handling bad punctuations and accurate sentence tokenization. Each sentence is tagged with their rhetorical roles. We also annotate each judgement with several attributes like date, names of the plaintiffs, defendants and the people representing them, judges who delivered the judgement, acts/statutes that are cited and the most common citations used to refer the judgement. Further, we propose an automatic labelling technique for identifying sentences which have \emph{summary worthy} information. We demonstrate that this auto-labeled data can be used effectively to train a weakly supervised sentence extractor with high accuracy. Some possible applications of this dataset besides legal document summarization can be in retrieval, citation analysis and prediction of decisions by a particular judge.
\end{abstract}

%% file: sections/keywords.tex
\begin{keyword}
Text Summarization \sep Legal Informatics
\end{keyword}

%% file: sections/Introduction.tex
 \section{Introduction}
 
There is an increasing interest in using the advances of NLP and machine learning to benefit the legal community. Especially tasks like summarization, rhetorical role labelling and precedent retrieval are highly sought after by lawyers and paralegals. Recently the Supreme Court of India established an AI Department, which focuses on administrative efficiency and legal research \footnote{\url{ https://vidhilegalpolicy.in/research/responsible-ai-for-the-indian-justice-system-a-strategy-paper/}}. It is clearly evident that AI integration in the Indian legal domain is no longer a futuristic endeavour. However, there are few publicly available large scale datasets of legal documents which is a necessity for any serious attempt at legal analytics research. In this work we try to mitigate that gap by proposing a new dataset consisting of over 10,000 judgements and corresponding summaries. While the data was already available in public domain, it was unstructured and noisy. We propose a pre-processing pipeline to curate the dataset and make it more usable.

Unlike the most commonly used datasets, like news articles, Wikipedia pages or social media content, the structure of legal text is much more richer. Very long and complex sentences with several abbreviations, named entities and citations are common in legal documents. The also have a variety of distinct units depending on the nature of information. For instance, statures are divided into sections, articles and paragraphs, while regulations can be in form of sections and sub-sections. This is country specific and varies considerably. As a result even the most basic pre-processing steps, like sentence and word tokenization, or entity recognition, do not work out of the box for legal texts. Further the legal vocabulary is highly technical and distinct from regular text, and is full of abbreviations which makes sentence tokenization even more difficult. This is commonly observed across a lot of works \cite{DBLP:conf/ecir/BhattacharyaHRP19, saravanan2006improving, sanchez2019sentence}.  As a result, although several tens of thousands of these court judgements are publicly available, their utility is limited until they can be properly processed. This work aims to bridge that gap, by providing a collection of pre-processed and annotated documents, that can be used for several downstream tasks. To the best of our knowledge, this is the largest publicly available annotated corpora of Indian legal text.

The proposed dataset consists of 10,764 supreme court judgements pre-processed by normalizing abbreviations, handling spell variations in named entities and tokenizing into sentences. Additionally, we identify other useful meta-information about the case, like date of judgement, involved parties, names of the judges, citations, etc. We also label each sentence with their rhetorical roles using the trained model made available by \cite{DBLP:conf/jurix/BhattacharyaPG019} and pseudo-relevance score which indicates whether or not that sentence is summary worthy. We demonstrate the use of this dataset in legal summarization using weakly supervised neural summarization. We conclude that neural approaches comfortably outperform strong baseline techniques. We also conclude that our approach, is effective on different time frames and different sub-domains. While we demonstrate the use on a summarization take, we believe the proposed dataset would be useful to several other tasks, including rhetorical role labelling, precedent retrieval and citation analysis as well. We plan to make the entire preprocessing and annotation pipeline publicly available and provide the required tools for loading and using this dataset. Some supplementary material including sample documents is made available in our github repository\footnote{\url{https://github.com/mtkh0602/LegalSummarization}}.

%% file: sections/Related_Works.tex
\section{Related Work}
A lot of progress has been made in legal text processing in past few years. This includes rhetorical role labelling\cite{saravanan2008automatic}\cite{DBLP:conf/jurix/BhattacharyaPG019}\cite{hachey2005sequence}, argument mining\cite{moens2007automatic}, legal text summarization\cite{moens2007summarizing}\cite{Mehta2019}\cite{DBLP:conf/ecir/BhattacharyaHRP19}, contract analysis, fact search/identification, etc. These works also generate several benchmark datasets, a few of which are publicly available. However, majority of the publicly available datasets are of small size. For example, the text summarization system for canadian case documents\cite{farzindar2004letsum} uses 10 document-summary pairs for evaluation. Likewise \cite{saravanan2006improving} uses 50 annotated judgements for evaluating graphical models of text summarization. While these datasets are not limited to English and several others like TEMIS in Italian\cite{venturi2012design} and Lexia in German\cite{waltl2016lexia} are available, most works are focused on legal documents from developed countries such as the UK, USA, Australia\ and Canada\cite{farzindar2004letsum}\cite{nejadgholi2017semi}. Few such corpora related to the Indian legal domain exist. In contrast to the existing datasets where the case documents are quite structured, the Indian case documents are a lot more noisy and not well structured. Indian case law reports do not usually contain any section headings and do not follow a certain structure, as compared to judgement from other countries like Austrilia, Canada, USA, making the summarization task more challenging.\cite{DBLP:conf/ecir/BhattacharyaHRP19}.

Despite this, there are several works that look at different aspects of legal text analysis in India like rhetorical role labelling\cite{DBLP:conf/jurix/BhattacharyaPG019}, semantic search\cite{DBLP:conf/fire/MorePPP19},\cite{DBLP:conf/fire/ZhaoNLHKHH19} or legal document summarization\cite{DBLP:conf/ecir/BhattacharyaHRP19, saravanan2006improving, DBLP:conf/icail/BhattacharyaPR021}. But the datasets used in these works again are either not public or relatively much smaller in size. The shared task on Artificial Intelligence for Legal Assistance\cite{DBLP:conf/fire/BhattacharyaG0019a, DBLP:conf/fire/Bhattacharya00G20} provides a publicly available dataset for searching precedence and statutes. It consists of about 3000 precedent cases, about 200 statutes and 50 queries (hypothetical scenarios). The task is to find the precedents and statutes relevant to the scenario. Another recent dataset related to rhetorical role labelling is provided by \cite{DBLP:conf/jurix/BhattacharyaPG019}. They provide 50 judgements tagged with 7 different rhetorical roles. A text summarization dataset used in \cite{DBLP:conf/ecir/BhattacharyaHRP19} is somewhat similar to what is proposed in the current work. The dataset consists of around 20,000 judgement and summary pairs from the supreme court of India, but it is not publicly available. Moreover, as mentioned in the paper, that dataset is not structured or annotated. In contrast, the proposed collection is both, much larger than most previous collections and is annotated with a lot of meta-information. 

Text summarization is a technique which refers to selecting the most important portions of original text and generating coherent summary out of it. The motivation for text summarization in legal domain is to help in navigating the enormous amount of legal documents produced across the world by the numerous legal institutions.  In India itself, there are 25 High Courts and 672 District Courts which publish the legal reports publicly. Such a system will also help in speeding up several cases that are pending in Indian courts [as of 2019, 87.5 percent, District and Subordinate courts] \footnote{\url{https://www.indiabudget.gov.in/budget2019-20/economicsurvey/doc/echapter.pdf}}. Since legal notes are long documents, legal institutions engage legal experts to produce headnotes which is known as summary. But such a task is labour-intensive, time-intensive as well as quite expensive. Therefore, Automatic summarization of legal documents can significantly help legal practitioners. Many summarization algorithms have been proposed till date, both for general text documents and a few specifically targeted to summarizing legal documents of various countries. 

For summarising Canadian case judgments, a method was proposed Letsum\cite{farzindar2004letsum} which determines the thematic structure of a judgment into four themes namely, introduction, context, juridical analysis, and conclusion. Relevant sentences are identified for each of the themes, and according to predefined percentages for each theme, the sentences are concatenated to form the summary. A similar approach was used in \cite{DBLP:conf/icail/BhattacharyaPR021} for summarizing Indian court judgements. In the case of Australian case judgments, the CaseSummarizer tool\cite{polsley-etal-2016-casesummarizer} was proposed which takes into account word frequencies and domain-specific information in the form of abbreviations, legal entities, etc. The legal entities like involved parties are recognized from section headings and also by identifying part of speech tags. Using these features, scores are determined for sentences, and depending on the threshold set by the user, summaries are generated. Another approach for summarization of Australian Case Judgements, HAUSS \cite{Galgani2014HAUSSIB} which combines various methods into one approach. Along with that many structural attributes are acquired at term and sentence levels. Based on which rules are created for the extraction of relevant sentences. For  UK case judgments \cite{Hachey2007ExtractiveSO} uses structural information along with manually annotated rhetorical roles for extractive summarization. Salomon \cite{Uyttendaele2004SalomonAA} is a legal summarization tool which was developed for Belgian criminal cases. It also takes into account structure of the judgement. Whole document is divided into segments, each having specific information and property. For example, alleged offences and opinions of court, both having irrelevant info, while verdict of court is to the point. The segments having irrelevant information are summarized by using clustering algorithms. All the legal tools mentioned are designed considering the format of the judgements published in respective countries. Therefore, these methods will not show good results for legal documents of another country as shown in \cite{DBLP:conf/ecir/BhattacharyaHRP19}.

One of the first works in text summarization for Indian judgments was by \cite{saravanan2008automatic} where conditional random field (CRF) based automatic text summarization is proposed. The authors have segmented the document into seven rhetorical roles. Then, various features are used for identification of labels. They presented their results on 200 Kerela High Court Judgements of which 50 were hand-annotated. In \cite{Kanapala2019SummarizationOL} each sentence of the judgement is ranked based on by optimizing fitness function which has sentence length, sentence position, degree of similarity, TF–ISF, legal keywords, etc. as their parameters. The proposed approach outperforms classical unsupervised algorithms. The dataset consisted of 1000 supreme court judgements. In \cite{DBLP:conf/ecir/BhattacharyaHRP19}, a systematic comparative study is performed on 17000+ Indian supreme court judgements, using algorithms such as LexRank\cite{Erkan2004LexRankGL}, LSA\cite{Steinberger2004UsingLS}, DSDR\cite{dsdr}, LetSum\cite{farzindar2004letsum}, Casesummarizer\cite{polsley-etal-2016-casesummarizer}, Graphical Model\cite{saravanan2006improving}, Neural Extractive Summarizer etc. 

All the approaches discussed above are extractive in nature, where most important sentences are identified and concatenated to form a summary. For, single-document extractive summarization, sentence representations of the documents are created and ranks sentences using machine learning algorithms. For example, SVM and Naive Bayes\cite{extractiveSumSVMNB} and Hidden Markov Model\cite{textSumHMM}. As the amount of labeled dataset increased, deep learning models started being used. In \cite{cao-etal-2015-learning}, CNN and LSTM are were used for creating sentence representations. In \cite{Cheng2016NeuralSB}, both CNN and LSTM are used for ranking the sentences. In \cite{Nallapati2017SummaRuNNerAR}, LSTM and GRU were used in a hierarchical manner for generating sentence representations. With onset state-of-the-art pre-trained language models several systems that used transformer like BERT\cite{Devlin2019BERTPO}, Roberta\cite{Liu2019RoBERTaAR} etc, were proposed for summarization \cite{Liu2019TextSW, Zhang2019HIBERTDL}. Recently, Google has also released \cite{Zaheer2020BigBT} for summarization of Long Documents. 

In case Unsupervised Extractive Summarisation, TextRank and LexRank were the prominent works. But, lately due increase in amount of text data published daily, and less labelled data many new approaches are coming up. The work by \cite{West2019BottleSumUA} proposes a novel approach for unsupervised sentence summarization by mapping the Information Bottleneck principle to a conditional language modelling objective. In \cite{zhou-rush-2019-simple}, two language models are used, and shows that by using a product-of-experts criteria is enough for maintaining continuous contextual matching while maintaining output fluency. In \cite{zheng-lapata-2019-sentence}, graph-based ranking algorithm is modified by computing node centrality in two ways,  employing BERT\cite{Devlin2019BERTPO} and graphs with directed edges arguing that the contribution of any two nodes to their respective centrality is influenced by their relative position in a document.

%% file: sections/dataset.tex
\section{Dataset Details}
\begin{figure}
    \centering
    \includegraphics[width=12cm]{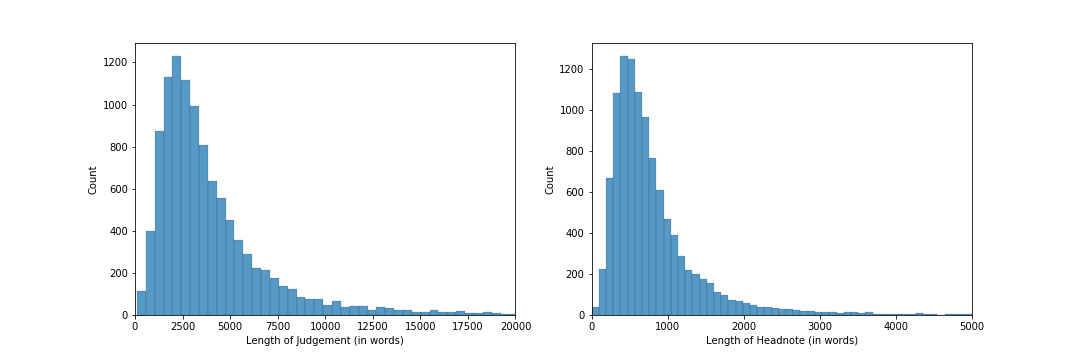}
    \caption{Frequency Distribution for Judgement and Headnote Lengths}
    \label{fig:freq}
\end{figure}

The dataset contains 10,764 judgements delivered by the supreme court of India which are publicly available\footnote{\url{www.liiofindia.org/in/cases/cen/INSC/}}. Table \ref{table1} shows some word and sentence sentence level statistics of the judgements and headnotes. Fig. \ref{fig:freq} shows the frequency distribution for length of judgements and headnotes in words. In this section we list out the annotation schema as well as the preprocessing steps used in creating this dataset. The end result is a structured json file for each judgement-headnote.\footnote{Complete json schema available at: \url{https://github.com/mtkh0602/LegalSummarization}}.

While most of the mentioned fields are annotated with high accuracy ($>$98$\%$), there are two exceptions. In case of identifying the people related to the case, there are instances where the information is not present at all, or where it is present but we identify it incorrectly or partially. The reason the latter scenario are use of incorrect punctuations and incorrect spellings. We estimate the accuracy of this field to be around 70\%. Add Precision Recall observation and Reasons for not proper identification and also example. The accuracy of rhetorical roles for sentences is limited by the original model used from \cite{DBLP:conf/jurix/BhattacharyaPG019} and estimated to be around 50$\%$ for our dataset.

\subsection{Preprocessing}
	The pre-processing steps consist mainly of normalizing abbreviations, identifying bad punctuations inserted by typing error(e.g.'appel- lant','consti- tuency', should have been a single words) and removing the extra spaces in sentences which are then used to improve sentence tokenization.

	\paragraph{\textbf{Abbreviation Normalization}}\label{acronyms} We identify a list of 31 most common abbreviations used in legal texts using regular expressions and frequency analysis. During pre-processing these abbreviations are expanded to their root form which serves two purposes. Apart from avoiding unexpected sentence breaks, removing unwanted punctuations, extra spaces between the words, expanding abbreviations also helps in normalizing the usage of those words. Several judgements use multiple abbreviations as well as their expanded form interchangeably (e.g. no. and num. for number, cls. and cl. for clause, etc). Expanding these reduces the vocabulary size, and the resulting consistency can be quite beneficial for down stream tasks. A complete list can be found in supplementary material. However, this list only consists of the most frequent abbreviations and is not exhaustive. There are several other, less frequent, abbreviations which are too numerous to handle explicitly. While we do identify them using the rules below, we do not expand them, but instead use that information to prevent incorrect sentence tokenization.
	
	We define a valid abbreviation the rules below, which were formed empirically.
	\begin{enumerate}
		\item[(a)] has four or fewer characters, ends with a full stop and has a document frequency of more than 20.
		\item[(b)] has a period between two characters (e.g. S.S.C)
	\end{enumerate}
	
	\paragraph{\textbf{Sentence Tokenization}} 
	After normalizing the 31 known abbreviations, we tokenize the text using the NLTK sentence tokenizer\footnote{\url{www.nltk.org}}. Next we post-process the output using a set of rules, which we define by observing the most frequent errors in original tokenization. We merge a sentence with the one immediately following it if the sentence ends in a valid acronym, as defined above or it ends in a number followed by a sentence that does not begin with capital letter. The latter is to handle a common problem where sections and clauses are usually referred to as \emph{'sec. 3.' or 'cls. 4.'} Unlike \cite{sanchez2019sentence} who find acronym handling to be ineffective in sentence boundary detection, in our case this preprocessing leads to correct sentence tokenization over 98$\%$ of the times.
	\input{sections/Eg_Tokenizer}

	Next we describe the annotation schema for the dataset. Each judgement and its associated headnote are converted into a json format with the fields described below. Apart from identifying sentences, we annotate several other fields, which can be potentially useful for summarization as well as several other tasks.

\subsection{Annotation Schema}

\paragraph{\textbf{Case Name and Judgement Date}} Case name is of the format \textbf{\textit{ABC and others vs. XYZ and others}}, where ABC are the plaintiffs and XYZ are the defendants. Date includes day, month and year of judgement. 

\paragraph{\textbf{Citations}}
Searching cases by standardized citations or using them to refer other judgements is a common practice. Every judgement delivered by the supreme court of India is indexed by several journals and databases, which are referred by legal professionals to find relevant past judgements. Each such source comes with its own citation format. In this dataset, we identify the citations from three most common sources, INSC (Indian Supreme court), AIR (All India Reporter) and SCR (Supreme court reports). In some cases citation information is available for other, less popular sources, besides the ones mentioned above which are clubbed under "Other citations" field. This field can be further used for citation analysis and also can be used for finding similar judgments. It can also be used for citation based approaches leverage other documents to summarize a target document. For a target document, they use the catchphrases of the documents cited by the target document (citphrases) and the citation sentences of documents that cite the target document (citances)\cite{galgani2012citation}. 

\paragraph{\textbf{Acts}} The field lists out the statutory acts referred to in the case. It includes a unique act id and the corresponding act text. These form an important part of the judgement and are usually included in the headnote. It is also particularly useful for finding precedents which relied on the same statutes while making an argument.

\paragraph{\textbf{Judges}} List of judges who delivered the verdict and additional fields  which indicates whether the judge was the chief justice or not, if judge deliverd the judgement or not. Often names of the judges are spelt differently across cases (e.g. P.K. Balasubramanyan and P. Balasubramanian refer to the same judge). To handle this we use a list of the justices who served in supreme court in the past, along with the years of their service, available from court website. We then match the name mentioned in the judgement to those in the list using levenshtein distance and date of the judgement. Such normalization makes it possible to filter the cases by judges which can have applications like predicting the outcome of case given a particular bench of judges. 

\paragraph{\textbf{Involved parties}} We identify the following parties involved in the judgements:
\begin{itemize}
	\item Plaintiffs, Defendants and Intervenors.
	\item People appearing for the plaintiffs, defendants and Intervenors. This is in form of list of tuples, the name of the person appearing and the person for whom she is appearing. Same person can appear for multiple plaintiffs, or multiple people can appear for the same plaintiff.
\end{itemize}

\paragraph{\textbf{Judgement Text}} 
This constitutes the actual judgement delivered by the court. This closely resembles the raw judgement text that is available publicly. The only change from the original judgement text is that we replace the most common abbreviations with their expanded form and removing erroneous punctuations. This also contains lot of meta information about the case, mentioned above, but as free text.

\paragraph{\textbf{Judgement Sentences}}
After preprocessing, we tokenize the judgement text into sentences using the steps mentioned in section \ref{acronyms}. 
We further annotate each judgement sentence with following information:

\begin{itemize}
	\item Pseudo-relevance of the sentence. This is a weak label that indicates whether a sentence is \emph{summary worthy} or not. A detailed discussion of this is presented in the next section. 
	\item Rhetorical role of the sentence as defined in in \cite{DBLP:conf/jurix/BhattacharyaPG019}. For this we train a Bi-LSTM CRF model using the code and training data provided by the authors.
\end{itemize}

\paragraph{\textbf{Headnote Text and Sentences}} The pre-processing and annotation process for headnote text and sentences is similar to that for the judgements, with one exception. The headnote sentence do not have any pseudo relevance score associated with them.

\subsection{Results and Analysis for Entities Identification}

In this section we report the results for entity identification mentioned in Annotation Schema.  We estimate the accuracy of all fields based on manual inspection of a subset of 50 randomly selected Supreme Court Judgements.

Case name, Judgement Date, Citations, Acts, are extracted from the Info file, which is provided along with Judgement and Headnote. We were able extract these entities correctly in all 50 Judgements.

For Judges, We were able to find all the judge who presided the case. We were able to identify judges in 98$\%$ of cases. The errors occurs due to, wrong usage of punctuations\footnote{\url{https://indiankanoon.org/doc/254621/}}, wrong sentence tokenization, and in some cases, the judge's name is not present at regular position in the judgement \footnote{\url{https://indiankanoon.org/doc/608874/}}. For Plaintiffs and Defendants, we used the case name. eg. DHIYAN SINGH AND ANR V. JUGAL KISHORE AND ANR, here Dhiyan Singh is the plaintiff and Jugal Kishore is the defendant. 
In some cases, Intervenors are also the part of Judgement. Intervenors are not a party to an existing lawsuit but who makes himself or herself a party either by joining with the plaintiff or uniting with the defendant in resistance of the plain-tiff's claims.\footnote{\url{https://legal-dictionary.thefreedictionary.com/Intervenor}}
Currently, we are able to identify the primary plaintiff, defendants, and intervenors. For other secondary parties, information is provided in very unstructured manner which makes it difficult to identify them.
Attorneys for parties , i.e, plaintiff, defendants and intervenors have accuracies of 70$\%$ , 73$\%$ and 100$\%$ respectively. Confusion matrices for all 3 are given in following figures. 


For each judgement sentence we are having two attributes, i.e, Rhetorical Roles and Pseudo Relevance. For Rhetorical Roles, we are 70$\%$ accurate in classifying the sentences into seven broad classes. 


For finding the Pseudo relevance for the sentences of the judgement, we followed below mentioned steps. They are:

\begin{itemize}
    \item For each case, we compute TF-IDF vectors for all judgement and headnote sentences.
    \item This vectors are used for calculating cosine similarity between each judgement sentence and all the headnote sentences. This would give us a matrix of dimension (Nos. of Judgement Sentences X Nos. of Headnote Sentences). 
    \item We take the maximum similarity value for each judgement sentence, and if value is greater than set threshold value, then sentence is classified as relevant.
\end{itemize}

For finding the threshold value, we manually annotated relevant sentences for 50 SupremeCourt cases. We tried different threshold values starting from 0.2 to 0.35. We found max correlation between manually annotated sentences and pseudo relevant sentences of 0.861 for threshold value of 0.3. Following figures \ref{fig:DiffThres} shows the confusion matrix for different thresholds. We have reported the results for Summarization using 0.3 threshold value.

\begin{figure}
    \centering
    \includegraphics[width=12cm]{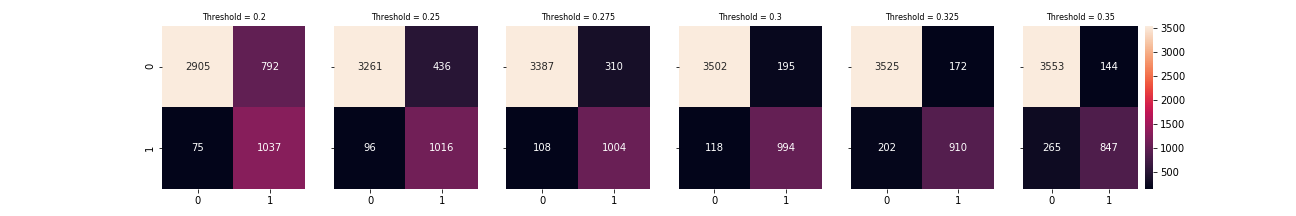}
    \caption{Confusion Matrix for various thresholds}
    \label{fig:DiffThres}
\end{figure}

%% file: sections/Eg_Tokenizer.tex
\begin{table}
{\resizebox{15cm}{!}{
\begin{tabular}{|l|l|}
\hline
 & Example of Sentence tokenization \footnote{\url{https://indiankanoon.org/doc/1857950/}}\\ \hline
 \begin{tabular}{l}
    \parbox[t]{2mm}{\multirow{8}{*}{\rotatebox[origin=c]{90}{ Original Legal Text }}}
  \end{tabular}
 & It was observed by Lord Atkin in Eshugbayi Eleko vs Officer Administering the Government of \\ & Nigeria C), that in accordance with British Jurispru dence no member of the executive can \\ & interfere with the liberty or property of a British subject except when he can support the legality \\ & of his act before a Court of justice. (1) [1924] 2 Irish Reports K. B. 104. (2) [1931] A. C. (62 at \\ &  670. 114 In The King vs The Secretary of State for Home Affairs(1),   Scrutton LJ. observed: \\ & "A man undoubtedly guilty of murder must yet be released if due forms of law have not \\ & been followed in  his conviction. " It seems very arguable that in the whole set-up of Part III \\ & of our Constitution these principles only remain guaranteed by article 21.
 \\ \hline
 \begin{tabular}{l}
 \parbox[t]{2mm}{\multirow{14}{*}{\rotatebox[origin=c]{90}{ NLTK Tokenization }}}
 \end{tabular}
 & 'It was observed by Lord Atkin in Eshugbayi Eleko vs Officer Administering the Government of \\ & Nigeria C), that in accordance with British Jurispru dence no member of the executive can interfere \\ & with the liberty or property of a British subject except when he can support the legality of his act \\ &  before a Court of justice.', \\ \cdashline{2-2}
    & '(1) [1924] 2 Irish Reports K. B.', \\ \cdashline{2-2}
    & '104.', \\\cdashline{2-2}
    & '(2) [1931] A. C. (62 at 670.', \\ \cdashline{2-2}
    & '114 In The King vs The Secretary of State for Home Affairs(1), Scrutton LJ.', \\ \cdashline{2-2}
    &'observed: "A man undoubtedly guilty of murder must \\ & yet be released if due forms of law have not been \\ & followed in his conviction. "', \\ \cdashline{2-2}
    &'It seems very arguable that in the whole set-up of Part III of our Constitution these principles only \\ & remain guaranteed by article 21.' \\ \hline
  \begin{tabular}{l}
 \parbox[t]{2mm}{\multirow{11}{*}{\rotatebox[origin=c]{90}{ LAWSUMM Tokenization }}}
 \end{tabular} & \textbf{'}It was observed by Lord Atkin in Eshugbayi Eleko vs Officer Administering the Government of \\ & Nigeria C), that in accordance with British Jurisprudence no member of the executive can interfere  \\ & with the liberty or property of a British subject except when he can support \\ & the legality  of his act before a Court of justice.\textbf{'}, \\ \cdashline{2-2}
   & \textbf{'}(1) [1924] 2 Irish Reports K. B. 104. (2) [1931] A. C. (62 at 670. 114 \\ & In The King vs The Secretary of State for Home Affairs(1), Scrutton \\ & LJ. observed: "A man undoubtedly guilty of murder must yet be released  \\ &  if due forms of law have not been followed in his conviction. "\textbf{'}, \\ \cdashline{2-2}
   & \textbf{'}It seems very arguable that in the whole set-up of Part III of our \\ &  Constitution these principles only remain guaranteed by article 21.\textbf{'}.
   \\ \hline
\end{tabular}
}}\end{table}

%% file: sections/experiments.tex
\section{Benchmark results for Legal Text Summarization}
In this section we report some late breaking results on legal summarization task. Use of the recent advances in neural summarization techniques in legal domain has been limited mainly due to lack of a publicly available annotated dataset. Given the complex nature and size of legal documents, which can easily run into dozens of pages, getting them manually annotated is prohibitively expensive. Instead we explore weak labelling techniques, similar to what is proposed in \cite{collins2017supervised} as an alternate to manual annotations. The headnotes are semi-abstractive and headnote sentences have considerable overlap with one or more judgement sentences. 


We exploit this fact, to create a weakly labelled training data. Each sentence in the judgement is considered to be \emph{summary worthy} if its cosine similarity with one or more headnote sentences is above the threshold of 0.3, set empirically as explained above.

\paragraph{\textbf{Data}} We use the 8648 Supreme Court Judgements as the training set. And the remaining Judgements were equally distributed for validation and testing data. We generate summaries and perform all evaluations on the test set. Unlike \cite{DBLP:conf/ecir/BhattacharyaHRP19} we have randomly distributed  judgment into 3 sets. A sentence classifier is then trained to identify \emph{summary worthy} sentences.

\paragraph{\textbf{Summary Length}} Some algorithms require the desired length of the summary to be given as an input. We tried different approaches for defining the summary length mentioned below: 

\begin{itemize}
    \item Using the mean ratio of number of words in a headnote to that in a judgement (\~23.4$\%$)
    \item Using the mean ratio of number of sentences in a headnote to that in a judgement (\~23.7$\%$)
    \item Mean headnote length in number of words (864 words) 
    \item Mean headnote length in terms of number of sentences (23 Sentences)
    \item Median headnote length in terms of number of words(610 words)
    \item Median headnote length in terms of number of sentences (20 Sentences)

\end{itemize}

\input{sections/model}

Threshold values were set on the classifier's output value for summary worthy sentences. We tried different threshold starting from 0.5 to 0.85 , with step size of 0.05. Best Rouge F1 scores of 0.619 were reported on 0.725.
 
\subsection{Results}
We compare our models to popular extractive summarization techniques like Lexrank\cite{Erkan2004LexRankGL}, LSA\cite{Steinberger2004UsingLS}, Greedy-KL\cite{haghighi-vanderwende-2009-exploring} and Sumbasic\cite{Sumbasic}. We also used the transformer based approach for extractive text summarization that uses BERT \cite{Liu2019TextSW} (implementation publicly available) for comparison.
Unlike \cite{DBLP:conf/ecir/BhattacharyaHRP19} we observe that neural network based approach outperforms these strong baseline techniques by a substantial margin. The results of this preliminary experiment are shown in table \ref{table2}. We report the standard ROUGE-F metrics for comparison\cite{lin-2004-rouge}. The baseline summaries are of 864 words, same as mean headnote length. For Neural summaries we do not choose the length explicitly and instead rely on the summarizer to classify sentences as important or not important. 
As evident, even a simple neural architecture comfortably outperforms strong baselines. This reinforces our hypothesis that weak labelling can be exploited for generating better summaries of legal documents. In future we would like to explore more suitable architecture and take advantage of other meta information like rhetorical roles, various entities, citations, etc. to improve the summaries.
\vspace{-6mm}
\begin{table}
\begin{center}

	\begin{tabular}{|c|c|c|}
		\hline
		             & Judge. & HeadN. \\ \hline
		 Mean Sents  &  134   & 20 \\
		Median Sents &   94   & 26 \\
		Min Sents & 3 & 0\\
		Max Sents & 3861 & 873 \\
		 Mean Words  &  4586  & 864 \\
		Median Words &  3194  & 610 \\
		Min Words & 103 & 0 \\
		Max Words & 139943 & 28408 \\ \hline
	\end{tabular}
\caption{Statistics of LAWSUMM\label{table1}}
\vspace{5mm}
\begin{tabular}{|c|c|c|c|}
	\hline
	             &         R1-F         &         R2-F         &         R4-F         \\ \hline
	  LexRank    &        0.542         &        0.286         &        0.134         \\
	    LSA      &        0.540         &        0.285         &        0.133         \\
	  GreedyKL   &        0.538         &        0.276         &        0.134         \\
	  Sumbasic   &        0.571         &        0.295         &        0.134         \\
	NN-Threshold & \textbf{0.619} & \textbf{0.408} & \textbf{0.261} \\ \hline
\end{tabular}
\caption{Benchmark Results\label{table2}}
\end{center}
\end{table}

%% file: sections/model.tex
\section{Model}
For this benchmark, we use a simple 2-layer Bidirectional-lstm neural network which has been proven to be good at sentence classification\cite{liu2019bidirectional}. We used hidden weights of 128 dimension with Dropout of 0.3. Forward and Backward Hidden Weights were concated and used as input for a linear layer with 2 output nodes for two classes namely, summary worthy and not worthy sentences. Word Embeddings of size 150 initialized using Xavier Normal Distribution. Pretrained embeddings, such as glove, fasttext, word2vec, etc., performed poorly as most of the legal vocabulary were missing. Maximum sentence length is set to 150 words, which is the mean length of the sentences in the Indian Supreme Court Judgements. Total vocabulary size used is 10000 words selected based on the frequency of words in the Indian Supreme Court Judgements. Training and Validation Batch size is set 32. We have used CrossEntropyLoss as the loss function with 4x bias towards summary-worthy sentences. This is done to reduce the effect due to Data Imbalance. Apart from this we also under sampled the dataset during training phase.

%% file: sections/Different_Approaches.tex
\subsection{Analysis for Different Summary Lengths}
In Approach 2 and 4 have high recall value, so many n-grams are in the generated which are not part of headnotes. Therefore, we decreased the nos. of words and sentences for the generated Summaries by using median nos. of words and sentences in Approach 3 and 5 which have comparative values for Recall, Precision and F1-Score. While comparing approaches which uses sentences as a unit of length versus approaches that uses words as a unit length, we can see that precision is higher for latter approaches. This is because, Sentences in the Indian Legal Judgements are longer as compared to normal text, with mean length of 150 words.

\begin{table}
\begin{center}
\begin{tabular}{|c|c|c|c|}
	\hline
	        &     R1-F1 Score    &    R1-Precision    &   R1-Recall    \\ 
	   \hline
	  Approach 1    &     0.577   &    0.580   &  0.619    \\
	  Approach 2    &     0.582   &    0.511   &  0.735    \\
	  Approach 3    &     0.571   &    0.578   &  0.612    \\
	  Approach 4    &     0.585   &    0.521   &  0.719    \\
	  Approach 5    &     0.582   &    0.558   &  0.660    \\
	  Approach 6    &     0.579   &    0.375   &  0.629    \\
    \hline
\end{tabular}
\caption{ Results for Different Approaches\label{table3}}
\end{center}
\end{table}

\paragraph{\textbf{Analysis for Different Legal Domains}}
Table 4 shows how generalizable the model is across the 5 different domains. The model gives good performance across all the domains, except for ‘Land $\&$ Property’, best performance on Intellectual property.

\begin{table}
\begin{center}
\begin{tabular}{|c|c|c|c|}
	\hline
	   Legal Domains     &     R1-F    &    R2-F    &   R4-F    \\ 
	   \hline
	  Land $\&$ Property    &     0.560   &    0.326   &  0.177    \\
	  Constitutional     &     0.574   &    0.311   &  0.150    \\
	  Labour $\&$ Industrial Law   &   0.676   &   0.417   &   0.268   \\
	  Intellectual Property    &    0.697   &   0.509   &   0.368   \\
	  Criminal    &    0.606   &    0.305     &   0.178   \\
    \hline
\end{tabular}
\caption{Statistics of LAWSUMM on different Legal Domains\label{table4}}
\end{center}
\end{table}

\paragraph{\textbf{Analysis for Summary in different time frames}}

As the years passes by, way of writing the judgements and language used also changes. So, we test our model for summarization from 1950 to 1993. Fig 3. shows a line plot of Rouge-F1 Scores vs. years. The graph shows that our model performs equally, except for [1950 - 1955] where the graph is dipping.

\begin{figure}
    \centering
    \includegraphics[width=12cm]{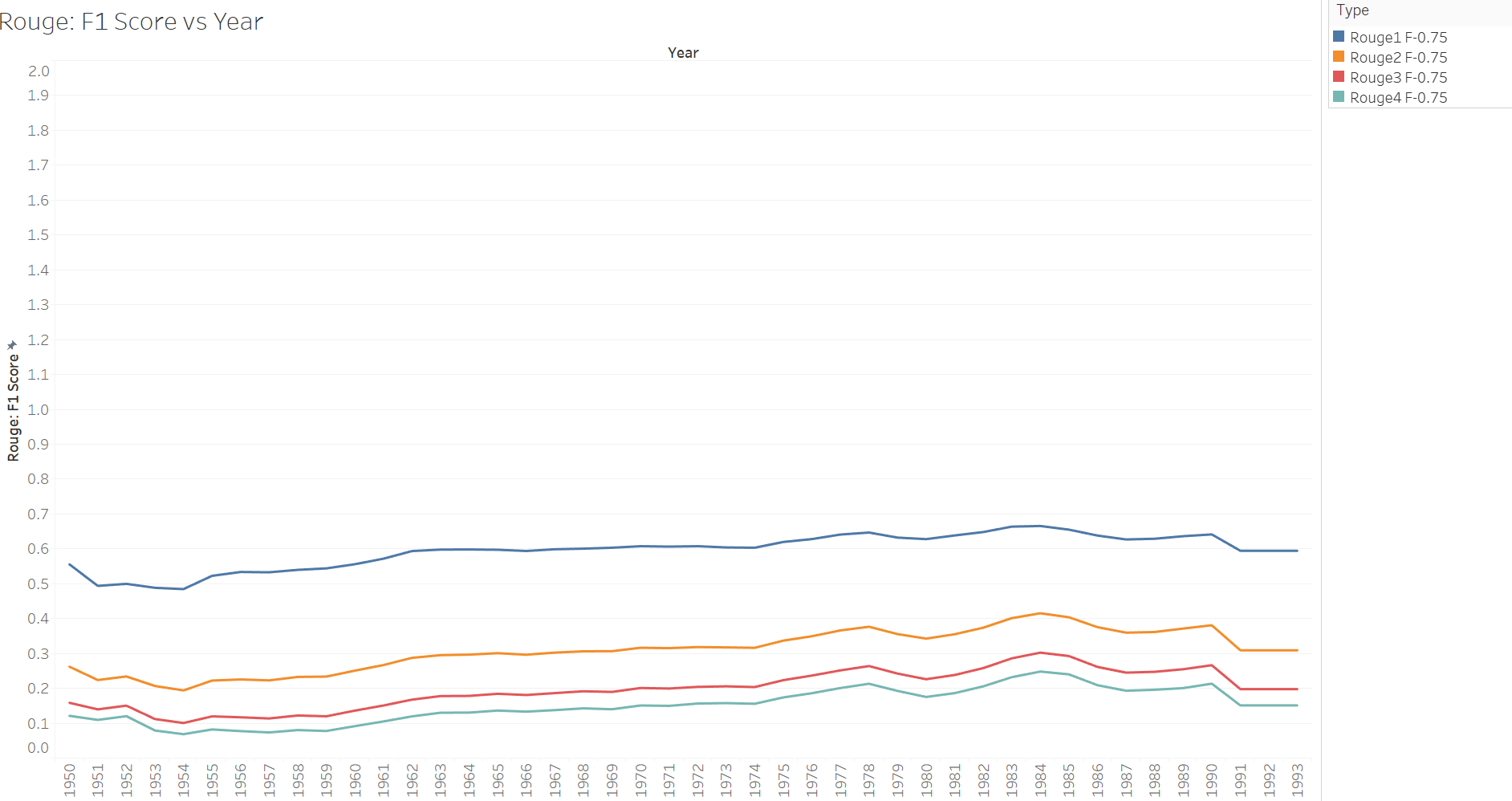}
    \caption{Rouge Scores vs.  Year}
\end{figure}

%% file: sections/conclusion.tex
\section{Conclusion} 
In this work we propose an annotated dataset of 10,764 judgements delivered by the supreme court of India, along with the associated handwritten summary called a headnote. We pre-process the documents to normalize abbreviations, named entities and tokenize it into sentences. We also annotate meta-information like names of people and judges associated with the case, date of judgement, citations of the case and statutory acts referred to in the judgement. Further we propose a weakly supervised approach for automatically summarizing these judgements. Some late breaking results show the effectiveness of our proposed weakly supervised approach which outperforms strong baseline techniques.

%% file: lawsumm.bbl
\begin{thebibliography}{45}
\expandafter\ifx\csname natexlab\endcsname\relax\def\natexlab#1{#1}\fi
\providecommand{\url}[1]{\texttt{#1}}
\providecommand{\href}[2]{#2}
\providecommand{\path}[1]{#1}
\providecommand{\DOIprefix}{doi:}
\providecommand{\ArXivprefix}{arXiv:}
\providecommand{\URLprefix}{URL: }
\providecommand{\Pubmedprefix}{pmid:}
\providecommand{\doi}[1]{\href{http://dx.doi.org/#1}{\path{#1}}}
\providecommand{\Pubmed}[1]{\href{pmid:#1}{\path{#1}}}
\providecommand{\bibinfo}[2]{#2}
\ifx\xfnm\relax \def\xfnm[#1]{\unskip,\space#1}\fi
\bibitem[{Bhattacharya et~al.(2019)Bhattacharya, Hiware, Rajgaria, Pochhi,
  Ghosh, and Ghosh}]{DBLP:conf/ecir/BhattacharyaHRP19}
\bibinfo{author}{P.~Bhattacharya}, \bibinfo{author}{K.~Hiware},
  \bibinfo{author}{S.~Rajgaria}, \bibinfo{author}{N.~Pochhi},
  \bibinfo{author}{K.~Ghosh}, \bibinfo{author}{S.~Ghosh},
\newblock \bibinfo{title}{A comparative study of summarization algorithms
  applied to legal case judgments},
\newblock in: \bibinfo{booktitle}{Advances in Information Retrieval - 41st
  European Conference on {IR} Research, {ECIR} 2019, Cologne, Germany, April
  14-18, 2019, Proceedings, Part {I}}, volume \bibinfo{volume}{11437} of
  \textit{\bibinfo{series}{Lecture Notes in Computer Science}},
  \bibinfo{publisher}{Springer}, \bibinfo{year}{2019}, pp.
  \bibinfo{pages}{413--428}.
\bibitem[{Saravanan et~al.(2006)Saravanan, Ravindran, and
  Raman}]{saravanan2006improving}
\bibinfo{author}{M.~Saravanan}, \bibinfo{author}{B.~Ravindran},
  \bibinfo{author}{S.~Raman},
\newblock \bibinfo{title}{Improving legal document summarization using
  graphical models},
\newblock \bibinfo{journal}{Frontiers in Artificial Intelligence and
  Applications} \bibinfo{volume}{152} (\bibinfo{year}{2006})
  \bibinfo{pages}{51}.
\bibitem[{Sanchez(2019)}]{sanchez2019sentence}
\bibinfo{author}{G.~Sanchez},
\newblock \bibinfo{title}{Sentence boundary detection in legal text},
\newblock in: \bibinfo{booktitle}{Proceedings of the Natural Legal Language
  Processing Workshop 2019}, \bibinfo{year}{2019}, pp. \bibinfo{pages}{31--38}.
\bibitem[{Bhattacharya et~al.(2019)Bhattacharya, Paul, Ghosh, Ghosh, and
  Wyner}]{DBLP:conf/jurix/BhattacharyaPG019}
\bibinfo{author}{P.~Bhattacharya}, \bibinfo{author}{S.~Paul},
  \bibinfo{author}{K.~Ghosh}, \bibinfo{author}{S.~Ghosh},
  \bibinfo{author}{A.~Wyner},
\newblock \bibinfo{title}{Identification of rhetorical roles of sentences in
  indian legal judgments},
\newblock in: \bibinfo{booktitle}{Legal Knowledge and Information Systems -
  {JURIX} 2019: The Thirty-second Annual Conference, Madrid, Spain, December
  11-13, 2019}, volume \bibinfo{volume}{322} of
  \textit{\bibinfo{series}{Frontiers in Artificial Intelligence and
  Applications}}, \bibinfo{publisher}{{IOS} Press}, \bibinfo{year}{2019}, pp.
  \bibinfo{pages}{3--12}.
\bibitem[{Saravanan et~al.(2008)Saravanan, Ravindran, and
  Raman}]{saravanan2008automatic}
\bibinfo{author}{M.~Saravanan}, \bibinfo{author}{B.~Ravindran},
  \bibinfo{author}{S.~Raman},
\newblock \bibinfo{title}{Automatic identification of rhetorical roles using
  conditional random fields for legal document summarization},
\newblock in: \bibinfo{booktitle}{Proceedings of the Third International Joint
  Conference on Natural Language Processing: Volume-{I}}, \bibinfo{year}{2008},
  pp. \bibinfo{pages}{481--490}. \URLprefix
  \url{https://aclanthology.org/I08-1063}.
\bibitem[{Hachey and Grover(2005)}]{hachey2005sequence}
\bibinfo{author}{B.~Hachey}, \bibinfo{author}{C.~Grover},
\newblock \bibinfo{title}{Sequence modelling for sentence classification in a
  legal summarisation system},
\newblock in: \bibinfo{booktitle}{Proceedings of the 2005 ACM Symposium on
  Applied Computing}, SAC '05, \bibinfo{publisher}{Association for Computing
  Machinery}, \bibinfo{address}{New York, NY, USA}, \bibinfo{year}{2005}, p.
  \bibinfo{pages}{292–296}. \URLprefix
  \url{https://doi.org/10.1145/1066677.1066746}.
  \DOIprefix\doi{10.1145/1066677.1066746}.
\bibitem[{Moens et~al.(2007)Moens, Boiy, Palau, and Reed}]{moens2007automatic}
\bibinfo{author}{M.-F. Moens}, \bibinfo{author}{E.~Boiy},
  \bibinfo{author}{R.~M. Palau}, \bibinfo{author}{C.~Reed},
\newblock \bibinfo{title}{Automatic detection of arguments in legal texts},
\newblock in: \bibinfo{booktitle}{Proceedings of the 11th International
  Conference on Artificial Intelligence and Law}, ICAIL '07,
  \bibinfo{publisher}{Association for Computing Machinery},
  \bibinfo{address}{New York, NY, USA}, \bibinfo{year}{2007}, p.
  \bibinfo{pages}{225–230}. \URLprefix
  \url{https://doi.org/10.1145/1276318.1276362}.
  \DOIprefix\doi{10.1145/1276318.1276362}.
\bibitem[{Moens(2007)}]{moens2007summarizing}
\bibinfo{author}{M.-F. Moens},
\newblock \bibinfo{title}{Summarizing court decisions},
\newblock \bibinfo{journal}{Information Processing and Management}
  \bibinfo{volume}{43} (\bibinfo{year}{2007}) \bibinfo{pages}{1748--1764}.
  \bibinfo{note}{Text Summarization}.
\bibitem[{Mehta and Majumder(2019)}]{Mehta2019}
\bibinfo{author}{P.~Mehta}, \bibinfo{author}{P.~Majumder},
  \bibinfo{title}{Domain-Specific Summarisation}, \bibinfo{publisher}{Springer
  Singapore}, \bibinfo{address}{Singapore}, \bibinfo{year}{2019}, pp.
  \bibinfo{pages}{35--48}. \URLprefix
  \url{https://doi.org/10.1007/978-981-13-8934-4_4}.
  \DOIprefix\doi{10.1007/978-981-13-8934-4_4}.
\bibitem[{Farzindar and Lapalme(2004)}]{farzindar2004letsum}
\bibinfo{author}{A.~Farzindar}, \bibinfo{author}{G.~Lapalme},
\newblock \bibinfo{title}{Letsum, an automatic legal text summarizing system},
\newblock \bibinfo{journal}{Legal knowledge and information systems, JURIX}
  (\bibinfo{year}{2004}) \bibinfo{pages}{11--18}.
\bibitem[{Venturi(2012)}]{venturi2012design}
\bibinfo{author}{G.~Venturi},
\newblock \bibinfo{title}{Design and development of temis: a syntactically and
  semantically annotated corpus of italian legislative texts},
\newblock in: \bibinfo{booktitle}{Proceedings of the Workshop on Semantic
  Processing of Legal Texts (SPLeT 2012)}, \bibinfo{year}{2012}, pp.
  \bibinfo{pages}{1--12}.
\bibitem[{Waltl et~al.(2016)Waltl, Matthes, Waltl, and Grass}]{waltl2016lexia}
\bibinfo{author}{B.~Waltl}, \bibinfo{author}{F.~Matthes},
  \bibinfo{author}{T.~Waltl}, \bibinfo{author}{T.~Grass},
\newblock \bibinfo{title}{Lexia: A data science environment for semantic
  analysis of german legal texts},
\newblock \bibinfo{journal}{Jusletter IT} \bibinfo{volume}{4}
  (\bibinfo{year}{2016}) \bibinfo{pages}{4--1}.
\bibitem[{Nejadgholi et~al.(2017)Nejadgholi, Bougueng, and
  Witherspoon}]{nejadgholi2017semi}
\bibinfo{author}{I.~Nejadgholi}, \bibinfo{author}{R.~Bougueng},
  \bibinfo{author}{S.~Witherspoon},
\newblock \bibinfo{title}{A semi-supervised training method for semantic search
  of legal facts in canadian immigration cases.},
\newblock in: \bibinfo{booktitle}{JURIX}, \bibinfo{year}{2017}, pp.
  \bibinfo{pages}{125--134}.
\bibitem[{More et~al.(2019)More, Patil, Palaskar, and
  Pawde}]{DBLP:conf/fire/MorePPP19}
\bibinfo{author}{R.~More}, \bibinfo{author}{J.~Patil},
  \bibinfo{author}{A.~Palaskar}, \bibinfo{author}{A.~Pawde},
\newblock \bibinfo{title}{Removing named entities to find precedent legal
  cases},
\newblock in: \bibinfo{booktitle}{Working Notes of {FIRE} 2019 - Forum for
  Information Retrieval Evaluation, Kolkata, India, December 12-15, 2019},
  volume \bibinfo{volume}{2517} of \textit{\bibinfo{series}{{CEUR} Workshop
  Proceedings}}, \bibinfo{publisher}{CEUR-WS.org}, \bibinfo{year}{2019}, pp.
  \bibinfo{pages}{13--18}.
\bibitem[{Zhao et~al.(2019)Zhao, Ning, Liu, Huang, Kong, Han, and
  Han}]{DBLP:conf/fire/ZhaoNLHKHH19}
\bibinfo{author}{Z.~Zhao}, \bibinfo{author}{H.~Ning}, \bibinfo{author}{L.~Liu},
  \bibinfo{author}{C.~Huang}, \bibinfo{author}{L.~Kong},
  \bibinfo{author}{Y.~Han}, \bibinfo{author}{Z.~Han},
\newblock \bibinfo{title}{Fire2019@aila: Legal information retrieval using
  improved {BM25}},
\newblock in: \bibinfo{booktitle}{Working Notes of {FIRE} 2019 - Forum for
  Information Retrieval Evaluation, Kolkata, India, December 12-15, 2019},
  volume \bibinfo{volume}{2517} of \textit{\bibinfo{series}{{CEUR} Workshop
  Proceedings}}, \bibinfo{publisher}{CEUR-WS.org}, \bibinfo{year}{2019}, pp.
  \bibinfo{pages}{40--45}.
\bibitem[{Bhattacharya et~al.(2021)Bhattacharya, Poddar, Rudra, Ghosh, and
  Ghosh}]{DBLP:conf/icail/BhattacharyaPR021}
\bibinfo{author}{P.~Bhattacharya}, \bibinfo{author}{S.~Poddar},
  \bibinfo{author}{K.~Rudra}, \bibinfo{author}{K.~Ghosh},
  \bibinfo{author}{S.~Ghosh},
\newblock \bibinfo{title}{Incorporating domain knowledge for extractive
  summarization of legal case documents},
\newblock in: \bibinfo{editor}{J.~Maranh{\~{a}}o}, \bibinfo{editor}{A.~Z.
  Wyner} (Eds.), \bibinfo{booktitle}{{ICAIL} '21: Eighteenth International
  Conference for Artificial Intelligence and Law, S{\~{a}}o Paulo Brazil, June
  21 - 25, 2021}, \bibinfo{publisher}{{ACM}}, \bibinfo{year}{2021}, pp.
  \bibinfo{pages}{22--31}. \URLprefix
  \url{https://doi.org/10.1145/3462757.3466092}.
  \DOIprefix\doi{10.1145/3462757.3466092}.
\bibitem[{Bhattacharya et~al.(2019)Bhattacharya, Ghosh, Ghosh, Pal, Mehta,
  Bhattacharya, and Majumder}]{DBLP:conf/fire/BhattacharyaG0019a}
\bibinfo{author}{P.~Bhattacharya}, \bibinfo{author}{K.~Ghosh},
  \bibinfo{author}{S.~Ghosh}, \bibinfo{author}{A.~Pal},
  \bibinfo{author}{P.~Mehta}, \bibinfo{author}{A.~Bhattacharya},
  \bibinfo{author}{P.~Majumder},
\newblock \bibinfo{title}{Overview of the {FIRE} 2019 {AILA} track: Artificial
  intelligence for legal assistance},
\newblock in: \bibinfo{booktitle}{Working Notes of {FIRE} 2019 - Forum for
  Information Retrieval Evaluation, Kolkata, India, December 12-15, 2019},
  volume \bibinfo{volume}{2517} of \textit{\bibinfo{series}{{CEUR} Workshop
  Proceedings}}, \bibinfo{publisher}{CEUR-WS.org}, \bibinfo{year}{2019}, pp.
  \bibinfo{pages}{1--12}.
\bibitem[{Bhattacharya et~al.(2020)Bhattacharya, Mehta, Ghosh, Ghosh, Pal,
  Bhattacharya, and Majumder}]{DBLP:conf/fire/Bhattacharya00G20}
\bibinfo{author}{P.~Bhattacharya}, \bibinfo{author}{P.~Mehta},
  \bibinfo{author}{K.~Ghosh}, \bibinfo{author}{S.~Ghosh},
  \bibinfo{author}{A.~Pal}, \bibinfo{author}{A.~Bhattacharya},
  \bibinfo{author}{P.~Majumder},
\newblock \bibinfo{title}{{FIRE} 2020 {AILA} track: Artificial intelligence for
  legal assistance},
\newblock in: \bibinfo{editor}{P.~Majumder}, \bibinfo{editor}{M.~Mitra},
  \bibinfo{editor}{S.~Gangopadhyay}, \bibinfo{editor}{P.~Mehta} (Eds.),
  \bibinfo{booktitle}{{FIRE} 2020: Forum for Information Retrieval Evaluation,
  Hyderabad, India, December 16-20, 2020}, \bibinfo{publisher}{{ACM}},
  \bibinfo{year}{2020}, pp. \bibinfo{pages}{1--3}. \URLprefix
  \url{https://doi.org/10.1145/3441501.3441510}.
  \DOIprefix\doi{10.1145/3441501.3441510}.
\bibitem[{Polsley et~al.(2016)Polsley, Jhunjhunwala, and
  Huang}]{polsley-etal-2016-casesummarizer}
\bibinfo{author}{S.~Polsley}, \bibinfo{author}{P.~Jhunjhunwala},
  \bibinfo{author}{R.~Huang},
\newblock \bibinfo{title}{{C}ase{S}ummarizer: A system for automated
  summarization of legal texts},
\newblock in: \bibinfo{booktitle}{Proceedings of {COLING} 2016, the 26th
  International Conference on Computational Linguistics: System
  Demonstrations}, \bibinfo{publisher}{The COLING 2016 Organizing Committee},
  \bibinfo{address}{Osaka, Japan}, \bibinfo{year}{2016}, pp.
  \bibinfo{pages}{258--262}. \URLprefix
  \url{https://aclanthology.org/C16-2054}.
\bibitem[{Galgani et~al.(2014)Galgani, Compton, and
  Hoffmann}]{Galgani2014HAUSSIB}
\bibinfo{author}{F.~Galgani}, \bibinfo{author}{P.~Compton},
  \bibinfo{author}{A.~Hoffmann},
\newblock \bibinfo{title}{Hauss: Incrementally building a summarizer combining
  multiple techniques},
\newblock \bibinfo{journal}{Int. J. Hum. Comput. Stud.} \bibinfo{volume}{72}
  (\bibinfo{year}{2014}) \bibinfo{pages}{584--605}.
\bibitem[{Hachey and Grover(2007)}]{Hachey2007ExtractiveSO}
\bibinfo{author}{B.~Hachey}, \bibinfo{author}{C.~Grover},
\newblock \bibinfo{title}{Extractive summarisation of legal texts},
\newblock \bibinfo{journal}{Artificial Intelligence and Law}
  \bibinfo{volume}{14} (\bibinfo{year}{2007}) \bibinfo{pages}{305--345}.
\bibitem[{Uyttendaele et~al.(2004)Uyttendaele, Moens, and
  Dumortier}]{Uyttendaele2004SalomonAA}
\bibinfo{author}{C.~Uyttendaele}, \bibinfo{author}{M.-F. Moens},
  \bibinfo{author}{J.~Dumortier},
\newblock \bibinfo{title}{Salomon: Automatic abstracting of legal cases for
  effective access to court decisions},
\newblock \bibinfo{journal}{Artificial Intelligence and Law}
  \bibinfo{volume}{6} (\bibinfo{year}{2004}) \bibinfo{pages}{59--79}.
\bibitem[{Kanapala et~al.(2019)Kanapala, Jannu, and
  Pamula}]{Kanapala2019SummarizationOL}
\bibinfo{author}{A.~Kanapala}, \bibinfo{author}{S.~Jannu},
  \bibinfo{author}{R.~Pamula},
\newblock \bibinfo{title}{Summarization of legal judgments using gravitational
  search algorithm},
\newblock \bibinfo{journal}{Neural Computing and Applications}
  (\bibinfo{year}{2019}) \bibinfo{pages}{1--9}.
\bibitem[{Erkan and Radev(2004)}]{Erkan2004LexRankGL}
\bibinfo{author}{G.~Erkan}, \bibinfo{author}{D.~R. Radev},
\newblock \bibinfo{title}{Lexrank: Graph-based lexical centrality as salience
  in text summarization},
\newblock \bibinfo{journal}{J. Artif. Intell. Res.} \bibinfo{volume}{22}
  (\bibinfo{year}{2004}) \bibinfo{pages}{457--479}.
\bibitem[{Steinberger and Jezek(2004)}]{Steinberger2004UsingLS}
\bibinfo{author}{J.~Steinberger}, \bibinfo{author}{K.~Jezek},
\newblock \bibinfo{title}{Using latent semantic analysis in text summarization
  and summary evaluation},
\newblock in: \bibinfo{booktitle}{Proceedings of the 5th International
  Conference on Information Systems Implementation and Modelling},
  \bibinfo{year}{2004}, pp. \bibinfo{pages}{93--100}.
\bibitem[{He et~al.(2012)He, Chen, Bu, Wang, Zhang, Cai, and He}]{dsdr}
\bibinfo{author}{Z.~He}, \bibinfo{author}{C.~Chen}, \bibinfo{author}{J.~Bu},
  \bibinfo{author}{C.~Wang}, \bibinfo{author}{L.~Zhang},
  \bibinfo{author}{D.~Cai}, \bibinfo{author}{X.~He},
\newblock \bibinfo{title}{Document summarization based on data reconstruction},
\newblock in: \bibinfo{booktitle}{Proceedings of the Twenty-Sixth AAAI
  Conference on Artificial Intelligence}, AAAI'12, \bibinfo{publisher}{AAAI
  Press}, \bibinfo{year}{2012}, p. \bibinfo{pages}{620–626}.
\bibitem[{Wong et~al.(2008)Wong, Wu, and Li}]{extractiveSumSVMNB}
\bibinfo{author}{K.-F. Wong}, \bibinfo{author}{M.~Wu}, \bibinfo{author}{W.~Li},
\newblock \bibinfo{title}{Extractive summarization using supervised and
  semi-supervised learning},
\newblock in: \bibinfo{booktitle}{Proceedings of the 22nd International
  Conference on Computational Linguistics - Volume 1}, COLING '08,
  \bibinfo{publisher}{Association for Computational Linguistics},
  \bibinfo{address}{USA}, \bibinfo{year}{2008}, p. \bibinfo{pages}{985–992}.
\bibitem[{Conroy and O'leary(2001)}]{textSumHMM}
\bibinfo{author}{J.~M. Conroy}, \bibinfo{author}{D.~P. O'leary},
\newblock \bibinfo{title}{Text summarization via hidden markov models},
\newblock in: \bibinfo{booktitle}{Proceedings of the 24th Annual International
  ACM SIGIR Conference on Research and Development in Information Retrieval},
  SIGIR '01, \bibinfo{publisher}{Association for Computing Machinery},
  \bibinfo{address}{New York, NY, USA}, \bibinfo{year}{2001}, p.
  \bibinfo{pages}{406–407}. \URLprefix
  \url{https://doi.org/10.1145/383952.384042}.
  \DOIprefix\doi{10.1145/383952.384042}.
\bibitem[{Cao et~al.(2015)Cao, Wei, Li, Li, Zhou, and
  Wang}]{cao-etal-2015-learning}
\bibinfo{author}{Z.~Cao}, \bibinfo{author}{F.~Wei}, \bibinfo{author}{S.~Li},
  \bibinfo{author}{W.~Li}, \bibinfo{author}{M.~Zhou},
  \bibinfo{author}{H.~Wang},
\newblock \bibinfo{title}{Learning summary prior representation for extractive
  summarization},
\newblock in: \bibinfo{booktitle}{Proceedings of the 53rd Annual Meeting of the
  Association for Computational Linguistics and the 7th International Joint
  Conference on Natural Language Processing (Volume 2: Short Papers)},
  \bibinfo{publisher}{Association for Computational Linguistics},
  \bibinfo{address}{Beijing, China}, \bibinfo{year}{2015}, pp.
  \bibinfo{pages}{829--833}. \URLprefix
  \url{https://aclanthology.org/P15-2136}. \DOIprefix\doi{10.3115/v1/P15-2136}.
\bibitem[{Cheng and Lapata(2016)}]{Cheng2016NeuralSB}
\bibinfo{author}{J.~Cheng}, \bibinfo{author}{M.~Lapata},
\newblock \bibinfo{title}{Neural summarization by extracting sentences and
  words},
\newblock \bibinfo{journal}{ArXiv} \bibinfo{volume}{abs/1603.07252}
  (\bibinfo{year}{2016}).
\bibitem[{Nallapati et~al.(2017)Nallapati, Zhai, and
  Zhou}]{Nallapati2017SummaRuNNerAR}
\bibinfo{author}{R.~Nallapati}, \bibinfo{author}{F.~Zhai},
  \bibinfo{author}{B.~Zhou},
\newblock \bibinfo{title}{Summarunner: A recurrent neural network based
  sequence model for extractive summarization of documents},
\newblock in: \bibinfo{booktitle}{Proceedings of the Thirty-First AAAI
  Conference on Artificial Intelligence}, AAAI'17, \bibinfo{publisher}{AAAI
  Press}, \bibinfo{year}{2017}, p. \bibinfo{pages}{3075–3081}.
\bibitem[{Devlin et~al.(2018)Devlin, Chang, Lee, and
  Toutanova}]{Devlin2019BERTPO}
\bibinfo{author}{J.~Devlin}, \bibinfo{author}{M.-W. Chang},
  \bibinfo{author}{K.~Lee}, \bibinfo{author}{K.~Toutanova},
\newblock \bibinfo{title}{Bert: Pre-training of deep bidirectional transformers
  for language understanding},
\newblock \bibinfo{journal}{arXiv preprint arXiv:1810.04805}
  (\bibinfo{year}{2018}).
\bibitem[{Liu et~al.(2019)Liu, Ott, Goyal, Du, Joshi, Chen, Levy, Lewis,
  Zettlemoyer, and Stoyanov}]{Liu2019RoBERTaAR}
\bibinfo{author}{Y.~Liu}, \bibinfo{author}{M.~Ott}, \bibinfo{author}{N.~Goyal},
  \bibinfo{author}{J.~Du}, \bibinfo{author}{M.~Joshi},
  \bibinfo{author}{D.~Chen}, \bibinfo{author}{O.~Levy},
  \bibinfo{author}{M.~Lewis}, \bibinfo{author}{L.~Zettlemoyer},
  \bibinfo{author}{V.~Stoyanov},
\newblock \bibinfo{title}{Roberta: A robustly optimized bert pretraining
  approach},
\newblock \bibinfo{journal}{ArXiv} \bibinfo{volume}{abs/1907.11692}
  (\bibinfo{year}{2019}).
\bibitem[{Liu and Lapata(2019)}]{Liu2019TextSW}
\bibinfo{author}{Y.~Liu}, \bibinfo{author}{M.~Lapata}, \bibinfo{title}{Text
  summarization with pretrained encoders}, \bibinfo{year}{2019}.
  \href{http://arxiv.org/abs/1908.08345}{\tt arXiv:1908.08345}.
\bibitem[{Zhang et~al.(2019)Zhang, Wei, and Zhou}]{Zhang2019HIBERTDL}
\bibinfo{author}{X.~Zhang}, \bibinfo{author}{F.~Wei},
  \bibinfo{author}{M.~Zhou},
\newblock \bibinfo{title}{Hibert: Document level pre-training of hierarchical
  bidirectional transformers for document summarization},
\newblock \bibinfo{journal}{ArXiv} \bibinfo{volume}{abs/1905.06566}
  (\bibinfo{year}{2019}).
\bibitem[{Zaheer et~al.(2020)Zaheer, Guruganesh, Dubey, Ainslie, Alberti,
  Onta{\~n}{\'o}n, Pham, Ravula, Wang, Yang, and Ahmed}]{Zaheer2020BigBT}
\bibinfo{author}{M.~Zaheer}, \bibinfo{author}{G.~Guruganesh},
  \bibinfo{author}{K.~A. Dubey}, \bibinfo{author}{J.~Ainslie},
  \bibinfo{author}{C.~Alberti}, \bibinfo{author}{S.~Onta{\~n}{\'o}n},
  \bibinfo{author}{P.~Pham}, \bibinfo{author}{A.~Ravula},
  \bibinfo{author}{Q.~Wang}, \bibinfo{author}{L.~Yang},
  \bibinfo{author}{A.~Ahmed},
\newblock \bibinfo{title}{Big bird: Transformers for longer sequences},
\newblock \bibinfo{journal}{ArXiv} \bibinfo{volume}{abs/2007.14062}
  (\bibinfo{year}{2020}).
\bibitem[{West et~al.(2019)West, Holtzman, Buys, and
  Choi}]{West2019BottleSumUA}
\bibinfo{author}{P.~West}, \bibinfo{author}{A.~Holtzman},
  \bibinfo{author}{J.~Buys}, \bibinfo{author}{Y.~Choi},
  \bibinfo{title}{Bottlesum: Unsupervised and self-supervised sentence
  summarization using the information bottleneck principle},
  \bibinfo{year}{2019}. \href{http://arxiv.org/abs/1909.07405}{\tt
  arXiv:1909.07405}.
\bibitem[{Zhou and Rush(2019)}]{zhou-rush-2019-simple}
\bibinfo{author}{J.~Zhou}, \bibinfo{author}{A.~Rush},
\newblock \bibinfo{title}{Simple unsupervised summarization by contextual
  matching},
\newblock in: \bibinfo{booktitle}{Proceedings of the 57th Annual Meeting of the
  Association for Computational Linguistics}, \bibinfo{publisher}{Association
  for Computational Linguistics}, \bibinfo{address}{Florence, Italy},
  \bibinfo{year}{2019}, pp. \bibinfo{pages}{5101--5106}. \URLprefix
  \url{https://aclanthology.org/P19-1503}.
  \DOIprefix\doi{10.18653/v1/P19-1503}.
\bibitem[{Zheng and Lapata(2019)}]{zheng-lapata-2019-sentence}
\bibinfo{author}{H.~Zheng}, \bibinfo{author}{M.~Lapata},
\newblock \bibinfo{title}{Sentence centrality revisited for unsupervised
  summarization},
\newblock in: \bibinfo{booktitle}{Proceedings of the 57th Annual Meeting of the
  Association for Computational Linguistics}, \bibinfo{publisher}{Association
  for Computational Linguistics}, \bibinfo{address}{Florence, Italy},
  \bibinfo{year}{2019}, pp. \bibinfo{pages}{6236--6247}. \URLprefix
  \url{https://aclanthology.org/P19-1628}.
  \DOIprefix\doi{10.18653/v1/P19-1628}.
\bibitem[{Galgani~F.(2012)}]{galgani2012citation}
\bibinfo{author}{H.~A. Galgani~F., Compton~P.},
\newblock \bibinfo{title}{Citation based summarisation of legal texts},
\newblock in: \bibinfo{booktitle}{PRICAI 2012: Trends in Artificial
  Intelligence}, \bibinfo{year}{2012}, pp. \bibinfo{pages}{40--52}.
\bibitem[{Collins et~al.(2017)Collins, Augenstein, and
  Riedel}]{collins2017supervised}
\bibinfo{author}{E.~Collins}, \bibinfo{author}{I.~Augenstein},
  \bibinfo{author}{S.~Riedel},
\newblock \bibinfo{title}{A supervised approach to extractive summarisation of
  scientific papers},
\newblock in: \bibinfo{booktitle}{Proceedings of the 21st Conference on
  Computational Natural Language Learning (CoNLL 2017)}, \bibinfo{year}{2017},
  pp. \bibinfo{pages}{195--205}.
\bibitem[{Liu and Guo(2019)}]{liu2019bidirectional}
\bibinfo{author}{G.~Liu}, \bibinfo{author}{J.~Guo},
\newblock \bibinfo{title}{Bidirectional lstm with attention mechanism and
  convolutional layer for text classification},
\newblock \bibinfo{journal}{Neurocomputing} \bibinfo{volume}{337}
  (\bibinfo{year}{2019}) \bibinfo{pages}{325--338}.
\bibitem[{Haghighi and Vanderwende(2009)}]{haghighi-vanderwende-2009-exploring}
\bibinfo{author}{A.~Haghighi}, \bibinfo{author}{L.~Vanderwende},
\newblock \bibinfo{title}{Exploring content models for multi-document
  summarization},
\newblock in: \bibinfo{booktitle}{Proceedings of Human Language Technologies:
  The 2009 Annual Conference of the North {A}merican Chapter of the Association
  for Computational Linguistics}, \bibinfo{publisher}{Association for
  Computational Linguistics}, \bibinfo{address}{Boulder, Colorado},
  \bibinfo{year}{2009}, pp. \bibinfo{pages}{362--370}. \URLprefix
  \url{https://aclanthology.org/N09-1041}.
\bibitem[{Vanderwende et~al.(2007)Vanderwende, Suzuki, Brockett, and
  Nenkova}]{Sumbasic}
\bibinfo{author}{L.~Vanderwende}, \bibinfo{author}{H.~Suzuki},
  \bibinfo{author}{C.~Brockett}, \bibinfo{author}{A.~Nenkova},
\newblock \bibinfo{title}{Beyond sumbasic: Task-focused summarization with
  sentence simplification and lexical expansion},
\newblock \bibinfo{journal}{Inf. Process. Manage.} \bibinfo{volume}{43}
  (\bibinfo{year}{2007}) \bibinfo{pages}{1606–1618}.
\bibitem[{Lin(2004)}]{lin-2004-rouge}
\bibinfo{author}{C.-Y. Lin},
\newblock \bibinfo{title}{{ROUGE}: A package for automatic evaluation of
  summaries},
\newblock in: \bibinfo{booktitle}{Text Summarization Branches Out},
  \bibinfo{publisher}{Association for Computational Linguistics},
  \bibinfo{address}{Barcelona, Spain}, \bibinfo{year}{2004}, pp.
  \bibinfo{pages}{74--81}. \URLprefix \url{https://aclanthology.org/W04-1013}.

\end{thebibliography}
